\documentclass[journal]{IEEEtai}

\usepackage[colorlinks,urlcolor=blue,linkcolor=blue,citecolor=blue]{hyperref}

\usepackage{color,array}

\usepackage{graphicx}
\usepackage{multirow}%
\usepackage{amsmath,amssymb,amsfonts}%
\usepackage{amsthm}%
\usepackage{mathrsfs}%
\usepackage{textcomp}%
\usepackage{manyfoot}%
\usepackage{booktabs}%
\usepackage{algorithm}%
\usepackage{algorithmicx}%
\usepackage{algpseudocode}%
\usepackage{listings}%

\usepackage{natbib}
\usepackage{amssymb}
\let\labelindent\relax
\usepackage{enumitem}
\usepackage{pdfpages}
\usepackage{array}
\usepackage{geometry}
\usepackage{multirow}
\usepackage{amssymb}
\usepackage{pifont}
\usepackage{xcolor}
\usepackage{tablefootnote}

\newcommand{\cmark}{\ding{51}}%
\newcommand{\xmark}{\ding{55}}%
\newcommand{\goalbox}[1]{\fbox{#1}}
\newcommand{\filledgoalbox}[1]{\colorbox{black}{\color{white}\textbf{#1}}}

\usepackage{caption}

\begin{document}

\title{XAI meets LLMs: A Survey of the Relation between Explainable AI and Large Language Models}

\author{
\thanks{  }
\thanks{}}

\author{
\IEEEauthorblockN{Erik Cambria}\\
\IEEEauthorblockA{\textit{School of Computer Science and Engineering} \\
\textit{ Nanyang Technological University},
Singapore \\
cambria@ntu.edu.sg}\\
\and
\IEEEauthorblockN{Lorenzo Malandri}\\
\IEEEauthorblockA{\textit{Dept. of Statistics and Quantitative Methods} \\
\textit{University of Milano-Bicocca},
Milan, Italy \\
lorenzo.malandri@unimib.it}\\
\and
\IEEEauthorblockN{Fabio Mercorio}\\
\IEEEauthorblockA{\textit{Dept. of Statistics and Quantitative Methods} \\
\textit{University of Milano-Bicocca},
Milan, Italy \\
fabio.mercorio@unimib.it}\\
\and
\IEEEauthorblockN{Navid Nobani}\\
\IEEEauthorblockA{\textit{Dept. of Statistics and Quantitative Methods} \\
\textit{University of Milano-Bicocca},
Milan, Italy \\
navid.nobani@unimib.it}\\
\and
\IEEEauthorblockN{Andrea Seveso}\\
\IEEEauthorblockA{\textit{Dept. of Statistics and Quantitative Methods} \\
\textit{University of Milano-Bicocca},
Milan, Italy \\
andrea.seveso@unimib.it}\\
}

\maketitle

\begin{abstract}
In this survey, we address the key challenges in Large Language Models (LLM) research, focusing on the importance of interpretability. Driven by increasing interest from AI and business sectors, we highlight the need for transparency in LLMs. We examine the dual paths in current LLM research and eXplainable Artificial Intelligence (XAI): enhancing performance through XAI and the emerging focus on model interpretability. Our paper advocates for a balanced approach that values interpretability equally with functional advancements. Recognizing the rapid development in LLM research, our survey includes both peer-reviewed and preprint (arXiv) papers, offering a comprehensive overview of XAI's role in LLM research. We conclude by urging the research community to advance both LLM and XAI fields together.
\end{abstract}



\section{Introduction}

\IEEEPARstart{T}{he} emergence of LLMs has significantly impacted Artificial Intelligence (AI), given their excellence in several Natural Language Processing (NLP) applications. Their versatility reduces the need for handcrafted features, enabling applications across various domains. Their heightened creativity in content generation and contextual understanding contributes to advancements in creative writing and conversational AI. 
Additionally, extensive pre-training on large amounts of data enables LLMs to exhibit strong generalisation capacities without further domain-specific data from the user~\cite{zhao2023survey,amican}.
For those reasons, LLMs are swiftly becoming mainstream tools, deeply integrated into many industry sectors, such as medicine (see, e.g.,~\cite{thirunavukarasu2023large}) and finance (see, e.g.,~\cite{wu2023bloomberggpt}), to name a few. 

However, their emergence also raises ethical concerns, necessitating ongoing efforts to address issues related to bias, misinformation, and responsible AI deployment. 
LLMs are a notoriously complex ``black-box" system. Their inner workings are opaque, and their intricate complexity makes their interpretation challenging~\cite{kaadoud2021explainable,camsev}. Such opaqueness can lead to the production of inappropriate content or misleading outputs~\cite{weidinger2021ethical}. Finally, lacking visibility on their training data can further hinder trust and accountability in critical applications~\cite{liu2023importance}.

In this context, XAI is a crucial bridge between complex LLM-based systems and human understanding of their behaviour. Developing XAI frameworks for LLMs is essential for building user trust, ensuring accountability and fostering a responsible and ethical use of those models.

In this article, we review and categorise current XAI for LLMs in a structured manner.
Emphasising the importance of clear and truthful explanations, as suggested by~\cite{sevastjanova2022beware}, this survey aims to guide future research towards enhancing LLMs' explainability and trustworthiness in practical applications.

\subsection{Contribution}
\label{sec:contribution}
The contribution of our work is threefold:

\begin{enumerate}[leftmargin=*, noitemsep, topsep=0pt]
  \item We introduce a novel categorisation framework for assessing the body of research concerning the explainability of LLMs. The framework provides a clear and organised overview of the state of the art.
  
  \item We conduct a comprehensive survey of peer-reviewed and preprint papers based on ArXiv and DBLP databases, going beyond using common research tools.
  
  \item We critically assess current practices, identifying research gaps and issues and articulating potential future research trajectories.
\end{enumerate}

\subsection{Research questions} \label{sec:RQ}
In this survey, we explore the coexistence of XAI methods with LLMs and how these two fields are merged.
Specifically, our investigation revolves around these key questions:
\begin{enumerate}[label=Q\arabic*]
  \item How are XAI techniques currently being integrated with LLMs?
  \item What are the emerging trends in converging LLMs with XAI methodologies?
  \item What are the gaps in the current related literature, and what areas require further research?
\end{enumerate}

\section{The Need for Explanations in LLMs}
\label{sec:need}
In XAI field, the intersection with LLMs presents unique challenges and opportunities. This survey paper aims to dissect these challenges, extending the dialogue beyond the conventional understanding of XAI's objective, which is to illuminate the inner mechanisms of opaque models for various stakeholders while avoiding the introduction of new uncertainties (See e.g., ~\cite{cambria2023survey, burkart2021survey}).





Despite their advancements, LLMs struggle with complexity and opacity, raising design, deployment and interpretation issues. Inspired by~\cite{weidinger2021ethical}, this paper categorises LLM challenges into user-visible and invisible ones.

\paragraph{Visible User Challenges}
\leftskip=0em
\textit{Directly perceivable challenges for users without specialised tools.}

\paragraph{Trust and Transparency}
\leftskip=1em
Trust issues arise in crucial domains, e.g., healthcare~\cite{mercorio2020exdil,gozzi2022xai,alimonda2022survey} or finance~\cite{xing2020financial,castelnovo2023leveraging,yeo2023comprehensive}, due to the opacity of black-box models, including LLMs.
XAI must offer transparent, ethically aligned explanations for wider acceptance, especially under stringent regulations that mandate explainability (e.g., EU's GDPR~\cite{novelli2024generative}).
This impacts regulatory compliance and public credibility, with examples in European skill intelligence projects requiring XAI for decision explanations \cite{DBLP:journals/inffus/MalandriMMNS22, malandri2024model, malandri2022contrastive, malandri2022good}. 

\paragraph{Misuse and Critical Thinking Impacts}
\leftskip=1em
LLMs' versatility risks misuse, such as content creation for harmful purposes and evading moderation \cite{shen2023anything}. Over-reliance on LLMs may also erode critical thinking and independent analysis, as seen in educational contexts (see, e.g.~\cite{abd2023large}).

\paragraph{Invisible User Challenges}
\leftskip=0em
\textit{Challenges requiring deeper model understanding.}

\paragraph{Ethical and Privacy Concerns}
\leftskip=1em
Ethical dilemmas from LLM use, such as fairness and hate speech issues, and privacy risks like sensitive data exposure, require proactive measures and ethical guidelines \cite{weidinger2021ethical,yan2023practical,salimi2023large}.

\paragraph{Inaccuracies and Hallucinations}
\leftskip=1em
LLMs can generate false information, posing risks in various sectors like education, journalism, and healthcare. Addressing these issues involves improving LLM accuracy, educating users, and developing fact-checking systems \cite{rawte2023survey, azaria2023internal}.

\section{Methodology}
\label{sec:criteria}Systematic Mapping Studies (SMSs) are comprehensive surveys that categorise and summarise a range of published works in a specific research area, identifying literature gaps, trends, and future research needs. They are especially useful in large or under-explored fields where a detailed Systematic Literature Review (SLR) may not be feasible.

SMS and SLR follow a three-phase method (planning, conducting, reporting) but differ in their approach, as SMSs address broader questions, cover a wider range of publications with a less detailed review, and aim to provide an overview of the research field. In contrast, SLRs focus on specific questions, thoroughly review fewer publications, and strive for precise, evidence-based outcomes~\cite{barn2017conducting}.


Following~\cite{martinez2023model}, we designed our SMS for XAI and LLMs, including peer-reviewed and preprint papers. The latter choice is because we believe in rapidly evolving fields like computer science, including preprints offering access to the latest research, essential for a comprehensive review~\cite{oikonomidi2020changes}. 

We followed these steps to structure our SMS: Section~\ref{sec:RQ} proposes and defines the research questions, Section~\ref{sec:retrieval} describes how the paper retrieval has been performed; Section~\ref{sec:selection} describes the paper selection process based on the defined criteria; Section~\ref{sec: false+} explains who we dealt with false positive results and finally in Section~\ref{sec:results} we describe the obtained results.

\subsection{Paper retrieval} \label{sec:retrieval}
\begin{figure*}[ht]
  \centering
  \includegraphics[width=1\textwidth]{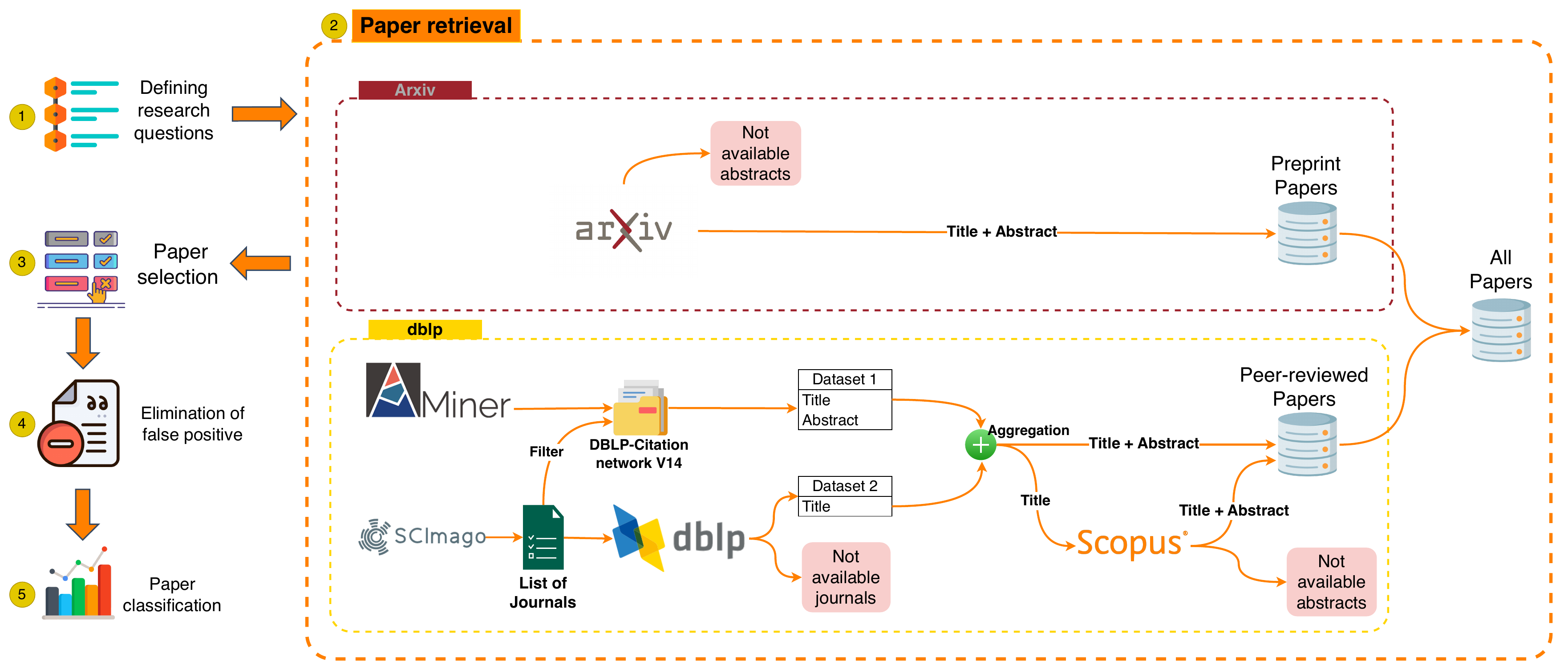}
  \caption{The process used for getting the papers related to our keywords, including the definition of research questions, paper retrieval, paper selection, elimination of false positives and classifying papers in the pre-defined categories.}
  \label{fig:process}
\end{figure*}

\paragraph{Overview} 
\leftskip=1em
Instead of utilising common scientific search engines such as Google Scholar, we employed a custom search methodology described in the following part. By scrutinising the titles and abstracts of the obtained papers, we conducted targeted searches using a predefined set of keywords pertinent to LLMs and XAI. This manual and deliberate search strategy was chosen to minimise the risk of overlooking relevant studies that automated search algorithms might miss and ensure our SMS dataset's accuracy and relevance. Through this rigorous process, we constructed a well-defined corpus of literature poised for in-depth analysis and review. Figure \ref{fig:process} provides an overview of this process.
\paragraph{Peer-reviewed papers} 
We initiated this step by identifying top-tier Q1 journals within the ``Artificial Intelligence" category of 2022 (last year available at the start of the study), providing us with 58 journals from which to draw relevant publications.

Subsequently, we utilised the XML dump\footnote{\url{https://dblp.org/xml/dblp.xml.gz}} from dblp computer science bibliography to get the titles of all papers published in the identified Q1 journals, except ten journals not covered by dblp. Once we gathered these paper titles, we proceeded to find their abstract. To do so, we initially used the last available citation network of AMiner\footnote{\url{https://originalfileserver.aminer.cn/misc/dblp\_v14.tar.gz}} but given that this dump lacks the majority of 2023 publications, we leveraged Scopus API, a detailed database of scientific abstracts and citations, to retrieve the missing abstracts corresponding to the amassed titles.

\paragraph{Pre-print papers}
We scraped all computer science papers presented in the Arxiv database from 2010 until October 2023, resulting in 548,711 papers. Consequently, we used the Arxiv API to get the abstracts of these papers.

\subsection{Paper selection} \label{sec:selection}
We employed a comprehensive set of keywords to filter the collected papers for relevance to LLMs and XAI. The search terms were carefully chosen to encompass the various terminologies and phrases commonly associated with each field.\footnote{ The keywords for XAI included: ['xai', 'explain', 'explanation', 'interpret', 'black box', 'black-box',
  'blackbox', 'transparent model understanding', 'feature importance',
  'accountable ai', 'ethical ai', 'trustworthy ai', 'fairness',
  'ai justification', 'causal inference', 'ai audit']

While for LLMs, the keywords are; ['llm', 'large language model', 'gpt-3', 'gpt-2', 'gpt3', 'gpt2', 'bert',
  'language model pre-training', 'fine-tuning language models',
  'generative pre-trained transformer', 
  'llama', ' bard', 'roberta', ' T5', 'xlnet', 'megatron', 'electra',
  'deberta', ' ernie', ' albert', ' bart', 'blenderbot',
  'open pre-trained transformer', 'mt-nlg', 'turing-nlg', 'pegasus',
  'gpt-3.5', 'gpt-4', 'gpt3.5', 'gpt4', ' cohere', 'claude', 'jurassic-1',
  'openllama', 'falcon', 'dolly', 'mpt', 'guanaco', 'bloom', ' alpaca',
  'openchatkit', 'gpt4all', 'flan-t5', 'orca']}

\noindent In our search, we applied a logical \texttt{OR} operator within the members of each list to capture any of the terms within a single category, and an \texttt{AND} operator was used between the two lists to ensure that only papers containing terms from both categories were retrieved for our analysis.

\subsection{Dealing with false positives} \label{sec: false+}
Upon completion of the initial retrieval phase, we identified a total of 1,030 manuscripts. Since some research keywords possess a broad meaning, for instance the words 'explain' and 'interpret' can be used in contexts different from the one of XAI, we retrieved few false positive papers, i.e., papers not dealing with both XAI and LLMs.
We excluded the false positives—publications that address only XAI or LLMs independently or none of them. To do so, we manually analysed the title and abstract of each paper. This meticulous vetting process resulted in 233 papers relevant to XAI and LLMs. 

Given that including all these papers in our survey was not feasible, we have selected the most relevant ones, based on their average number of citations per year. The whole research process resulted in 35 articles selected.






\section{Retrieval Results}
\label{sec:results}
We divide papers into two macro-categories of \textit{Applicaiton papers}, i.e., papers that somehow generated explanations, either towards explainability or to use them as a feature for another task, and \textit{Discussion papers}, i.e., papers that do not engage with explanation generation but address an issue or research gap regarding the explainable LLM models.

\subsection{Application Papers}
The first macro-category includes papers using LLMs in a methodology, tool, or task. Based on how LLMs are used, we further divide this category into two sub-categories as follows: \textit{"To explain"}, i.e., papers which try to explain how LLMs work and provide an insight into the opaque nature of these models. The second sub-category of papers called \textit{"As feature"}, uses the explanations and features generated by LLMs to improve the results of various tasks. The following parts discuss these sub-categories:

\begin{table*}[ht]

\scriptsize
\centering
\resizebox{2\columnwidth}{!}{%
\begin{tabular}{
>{\hspace{0pt}}m{0.29\linewidth}
>{\hspace{0pt}}m{0.04\linewidth}
>{\hspace{0pt}}m{0.04\linewidth}
>{\hspace{0pt}}m{0.08\linewidth}
>{\hspace{0pt}}m{0.2\linewidth}
>{\hspace{0pt}}m{0.11\linewidth}
>{\hspace{0pt}}c} 
\toprule
\textbf{Paper and Tool} & \textbf{Star} & \textbf{Fork} & \textbf{Update} & \textbf{Target} & \textbf{Agnostic} & \textbf{Goal} \\ 
\midrule
\cite{vig2019multiscale} \href{https://github.com/jessevig/bertviz}{BertViz} & 6.1k & 734 & 08/23 & Transformers & \cmark & \goalbox{C} \goalbox{E} \goalbox{IMP} \filledgoalbox{INT} \goalbox{R} \\
\cite{swamy2021interpreting} \href{https://github.com/epfml/interpret-lm-knowledge}{Experiments} & 19 & 2 & 05/22 & BERT-based LM & \xmark & \filledgoalbox{C} \goalbox{E} \goalbox{IMP} \goalbox{INT} \goalbox{R} \\
\cite{wu2021polyjuice} \href{https://github.com/tongshuangwu/polyjuice}{Polyjuice} & 90 & 16 & 08/22 & - & \cmark & \goalbox{C} \filledgoalbox{E} \goalbox{IMP} \goalbox{INT} \goalbox{R}\\
\cite{wang2022interpretability} \href{https://github.com/redwoodresearch/Easy-Transformer}{TransformerLens} & 48 & 161 & 01/23 & GPT2-small & \xmark & \goalbox{C} \filledgoalbox{E} \goalbox{IMP} \goalbox{INT} \goalbox{R}\\
\cite{menon2022visual} - & - & - & - & Vision-LM & \cmark & \goalbox{C} \goalbox{E} \filledgoalbox{IMP} \filledgoalbox{INT} \goalbox{R}\\
\cite{gao2023chatgpt} \href{https://github.com/ArrogantL/ChatGPT4CausalReasoning}{Experiments} & 17 & 0 & 10/23 & ChatGPT & \xmark & \goalbox{C} \goalbox{E} \goalbox{IMP} \goalbox{INT} \filledgoalbox{R} \\
\cite{pan2023unifying} - & - & - & - & LLMs & \cmark & \goalbox{C} \goalbox{E} \goalbox{IMP} \filledgoalbox{INT} \goalbox{R}\\
\cite{conmy2023towards} \href{https://github.com/ArthurConmy/Automatic-Circuit-Discovery}{ACDC} & 105 & 23 & 11/23 & Transformers & \cmark & \goalbox{C} \goalbox{E} \goalbox{IMP} \filledgoalbox{INT} \goalbox{R} \\
\cite{he2022rethinking} \href{https://github.com/HornHehhf/RR}{RR} & 38 & 2 & 02/23 & LLMs & \cmark & \goalbox{C} \filledgoalbox{E} \filledgoalbox{IMP} \goalbox{INT} \goalbox{R} \\
\cite{yoran2023answering} \href{https://github.com/oriyor/reasoning-on-cots}{MCR} & 71 & 9 & 01/24 & LLMs & \cmark & \goalbox{C} \goalbox{E} \goalbox{IMP} \goalbox{INT} \filledgoalbox{R} \\
\cite{sarti2023inseq} \href{https://github.com/inseq-team/inseq}{Inseq} & 250 & 26 & 01/24 & SeqGen models & \cmark & \goalbox{C} \goalbox{E} \goalbox{IMP} \filledgoalbox{INT} \goalbox{R} \\
\cite{wu2023interpretability} \href{https://github.com/frankaging/align-transformers}{Boundless DAS} & 0 & 17 & 01/24 & LLMs & \cmark & \goalbox{C} \filledgoalbox{E} \goalbox{IMP} \goalbox{INT} \goalbox{R} \\
\cite{li2023towards} \href{https://github.com/paihengxu/XICL}{XICL} & 1 & 3 & 11/23 & LLMs & \cmark & \goalbox{C} \filledgoalbox{E} \goalbox{IMP} \goalbox{INT} \goalbox{R} \\
\cite{chen2023lmexplainer} LMExplainer & - & - & - & LLMs & \cmark & \goalbox{C} \filledgoalbox{E} \goalbox{IMP} \goalbox{INT} \goalbox{R} \\
\cite{gao2023chat} Chat-REC & - & - & - & Rec. systems & \xmark & \goalbox{C} \filledgoalbox{E} \filledgoalbox{IMP} \goalbox{INT} \goalbox{R} \\
\midrule
\cite{zhang2022improved} \href{https://github.com/moqingyan/dsr-lm}{DSRLM} & 9 & 1 & 07/23 & LLMs & \cmark & \goalbox{C} \goalbox{E} \goalbox{IMP} \goalbox{INT} \filledgoalbox{R} \\
\cite{singh2023explaining} \href{https://github.com/csinva/imodelsX}{SASC} & 61 & 14 & 01/24 & LLMs & \cmark & \goalbox{C} \filledgoalbox{E} \goalbox{IMP} \goalbox{INT} \goalbox{R} \\
\cite{li2022explanations} - & - & - & - & LLMs & \cmark & \goalbox{C} \filledgoalbox{E} \filledgoalbox{IMP} \goalbox{INT} \goalbox{R} \\
\cite{ye2022unreliability} \href{https://github.com/xiye17/TextualExplInContext}{TextualExplInContext} & 11 & 2 & 02/23 & LLMs & \cmark & \goalbox{C} \filledgoalbox{E} \filledgoalbox{IMP} \goalbox{INT} \goalbox{R}\\
\cite{turpin2023language} \href{https://github.com/milesaturpin/cot-unfaithfulness}{Experiments} & 25 & 9 & 03/23 & LLMs & \cmark & \goalbox{C} \filledgoalbox{E} \goalbox{IMP} \goalbox{INT} \goalbox{R}\\
\cite{kang2023explainable} AutoSD & - & - & - & Debugging models & \xmark & \goalbox{C} \filledgoalbox{E} \filledgoalbox{IMP} \goalbox{INT} \goalbox{R} \\
\cite{krishna2023post} AMPLIFY & - & - & - & LLMs & \cmark & \goalbox{C} \goalbox{E} \filledgoalbox{IMP} \goalbox{INT} \goalbox{R} \\
\cite{yang2023language} \href{https://github.com/YueYANG1996/LaBo}{Labo} & 51 & 4 & 12/23 & CBM & \xmark & \goalbox{C} \goalbox{E} \filledgoalbox{IMP} \goalbox{INT} \goalbox{R} \\
\cite{bitton2023breaking} \href{https://whoops-benchmark.github.io/}{WHOOPS!} & - & - & - & LLMs & \cmark & \goalbox{C} \filledgoalbox{E} \goalbox{IMP} \goalbox{INT} \goalbox{R} \\
\cite{shi2023chatgraph} \href{https://github.com/sycny/ChatGraph}{Chatgraph} & 2 & 0 & 07/23 & LLMs & \cmark & \goalbox{C} \goalbox{E} \goalbox{IMP} \filledgoalbox{INT} \goalbox{R} \\
\bottomrule
\end{tabular}} 
\caption{Synthesis of recent application papers, summarising engagement indicators as of January 2024, update timelines, model specificity, and the overarching aims of each study. In the first section of the table, \textit{To Explain} papers are listed, and \textit{As Feature} works in the second. Stars, forks, and last updates are not reported (-) for papers lacking associated repositories. Target is the specific focus of the study, such as a particular type of language model. Agnostic indicates whether the study is model-agnostic or not. The goal represents the primary objective of each study: comparison of models (C), explanation (E), improvement (IMP), interpretability (INT), and reasoning (R).}

\label{tab:studies}
\end{table*}

\subsubsection{To Explain}
Most papers, i.e., 17 out of 35, fit into this sub-category, with most addressing the need for more interpretable and transparent LLMs. 

For instance, \citet{vig2019multiscale} introduces a visualisation tool for understanding the attention mechanism in Transformer models like BERT and GPT-2. Their proposed tool provides insights at multiple scales, from individual neurons to whole model layers, helping to detect model bias, locate relevant attention heads, and link neurons to model behaviour.

\cite{swamy2021interpreting} presents a methodology for interpreting the knowledge acquisition and linguistic skills of BERT-based language models by extracting knowledge graphs from these models at different stages of their training. Knowledge graphs are often used for explainable extrapolation reasoning~\cite{lin2023techs}.

\cite{wu2021polyjuice} propose Polyjuice, a general-purpose counterfactual generator.
This tool generates diverse, realistic counterfactuals by fine-tuning GPT-2 on multiple datasets, allowing for controlled perturbations regarding type and location.

\cite{wang2022interpretability} investigates the mechanistic interpretability of GPT-2 small, particularly its ability to identify indirect objects in sentences. 
 The study involves circuit analysis and reverse engineering of the model's computational graph, identifying specific attention heads and their roles in this task. 

\cite{menon2022visual} introduce a novel approach for visual classification using descriptions generated by LLMs.
 This method, which they term ``classification by description," involves using LLMs like GPT-3 to generate descriptive features of visual categories.
 These features are then used to classify images more accurately while providing more transparent results than traditional methods that rely solely on category names. 

\cite{gao2023chatgpt} examines ChatGPT's capabilities in causal reasoning using tasks like Event Causality Identification (ECI), Causal Discovery (CD), and Causal Explanation Generation (CEG).
The authors claim that while ChatGPT is effective as a causal explainer, it struggles with causal reasoning and often exhibits causal hallucinations.
The study also investigates the impact of In-Context Learning (ICL) and Chain-of-Thought (CoT) techniques, concluding that ChatGPT's causal reasoning ability is highly sensitive to the structure and wording of prompts.

\cite{pan2023unifying} is a framework that aims to enhance LLMs with explicit, structured knowledge from KGs, addressing issues like hallucinations and lack of interpretability. 
The paper outlines three main approaches: KG-enhanced LLMs, LLM-augmented KGs, and synergised LLMs with KGs. 
This unification improves the performance and explainability of AI systems in various applications.
 
\cite{conmy2023towards} focuses on automating a part of the mechanistic interpretability workflow in neural networks.
 Using algorithms like Automatic Circuit Discovery (ACDC), the authors automate the identification of sub-graphs in neural models that correspond to specific behaviours or functionalities. 

\cite{he2022rethinking} presents a novel post-processing approach for LLMs that leverages external knowledge to enhance the faithfulness of explanations and improve overall performance. 
This approach, called Rethinking with Retrieval, uses CoT prompting to generate reasoning paths refined with relevant external knowledge. The authors claim that their method significantly improves the performance of LLMs on complex reasoning tasks by producing more accurate and reliable explanations.

Multi-Chain Reasoning (MCR) introduced by~\cite{yoran2023answering} improves question-answering in LLMs by prompting them to meta-reason over multiple reasoning chains. 
This approach helps select relevant facts, mix information from different chains, and generate better explanations for the answers. 
The paper demonstrates MCR's superior performance over previous methods, especially in multi-hop question-answering.

Inseq~\cite{sarti2023inseq} is a Python library that facilitates interpretability analyses of sequence generation models.
The toolkit focuses on extracting model internals and feature importance scores, particularly for transformer architectures.
It centralises access to various feature attribution methods, intuitively representable with visualisations such as heatmaps~\cite{aminimehr2023tbexplain}, promoting fair and reproducible evaluations of sequence generation models. 

Boundless Distributed Alignment Search (Boundless DAS) introduced by~\cite{wu2023interpretability} is a method for identifying interpretable causal structures in LLMs. 
In their paper, the authors demonstrate that the Alpaca model, a 7B parameter LLM, solves numerical reasoning problems by implementing simple algorithms with interpretable boolean variables. 

\cite{li2023towards} investigate how various demonstrations influence ICL in LLMs by exploring the impact of contrastive input-label demonstration pairs, including label flipping, input perturbation, and adding complementary explanations.
The study employs saliency maps to qualitatively and quantitatively analyse how these demonstrations affect the predictions of LLMs. 

LMExplainer~\cite{chen2023lmexplainer} is a method for interpreting the decision-making processes of LMs.
This approach combines a knowledge graph and a graph attention neural network to explain the reasoning behind an LM's predictions.

\cite{gao2023chat} propose a novel recommendation system framework, Chat-REC, which integrates LLMs for generating more interactive and explainable recommendations.
The system converts user-profiles and interaction histories into prompts for LLMs, enhancing the recommendation process with the ICL capabilities of LLMs.

DSR-LM proposed by~\cite{zhang2022improved} is a framework combining differentiable symbolic reasoning with pre-trained language models. 
The authors claim their framework improves logical reasoning in language models through a symbolic module that performs deductive reasoning, enhancing accuracy on deductive reasoning tasks. 

\subsubsection{As Feature}
Papers in this sub-category do not directly aim to provide more transparent models or explain LLM-based models. Instead, they use LLMs to generate reasoning and descriptions, which are used as input to a secondary task. 

For instance,~\cite{li2022explanations} explore how LLMs' explanations can enhance the reasoning capabilities of smaller language models (SLMs). 
They introduce a multi-task learning framework where SLMs are trained with explanations from LLMs, leading to improved performance in reasoning tasks.

\cite{ye2022unreliability} evaluates the reliability of explanations generated by LLMs in few-shot learning scenarios. 
The authors claim that LLM explanations often do not significantly improve learning performance and can be factually unreliable by highlighting the potential misalignment between LLM reasoning and factual correctness in their explanations.

\cite{turpin2023language} investigates the reliability of CoT reasoning. The authors claim that while CoT can improve task performance, it can also systematically misrepresent the true reason behind a model's prediction. They demonstrate this through experiments showing how biasing features in model inputs, such as reordering multiple-choice options, can heavily influence CoT explanations without being acknowledged in the explanation itself.

\cite{kang2023explainable} introduce an approach for automating the debugging process called Automated Scientific Debugging (AutoSD). This approach leverages LLMs to generate hypotheses about bugs in code and uses debuggers to interact with the buggy code. This approach leads to automated conclusions and patch generation and provides clear explanations for the debugging decisions, potentially leading to more efficient and accurate decisions by developers.

\cite{krishna2023post} present a framework called Amplifying Model Performance by Leveraging In-Context Learning with Post Hoc Explanations (AMPLIFY), aiming to improve the performance of LLMs on complex reasoning and language understanding tasks by automating the generation of rationales.
It leverages post hoc explanation methods, which output attribution scores indicating the influence of each input feature on model predictions, to construct natural language rationales. These rationales provide corrective signals to LLMs.

\cite{yang2023language} introduces Language Guided Bottlenecks (LaBo), a method for constructing high-performance Concept Bottleneck Models (CBMs) without manual specification of concepts. LaBo leverages GPT-3 to generate factual sentences about categories, forming candidate concepts for CBMs. 
These concepts are then aligned with images using CLIP \cite{radford2021learning} to form a bottleneck layer. The method efficiently searches for bottlenecks using a submodular utility, focusing on discriminative and diverse information. 
The authors claim their method outperforms black box linear probes in few-shot classification tasks across 11 diverse datasets, showing comparable or better performance with more data. 

\cite{bitton2023breaking} introduces WHOOPS!, a new dataset and benchmark designed to test AI models' visual commonsense reasoning abilities.
The dataset comprises images intentionally defying commonsense, created using image generation tools like Midjourney.
The paper assesses AI models on tasks such as image captioning, cross-modal matching, visual question answering, and the challenging task of explanation generation, where models must identify and explain the unusualness of an image.
Results show that even advanced models like GPT3 and BLIP2 struggle with these tasks, highlighting a gap in AI's visual commonsense reasoning compared to human performance.

\subsection{Discussion Papers}
Unlike the Application papers, this category includes papers that target the argument of XAI through LLMs and vice versa but do not necessarily provide any specific methodology, framework or application.
This category, in turn, is divided into two subcategories of \textbf{Issues}, or works which mention a \textit{concern} and \textbf{Benchmark and Metrics}, which mainly focus on evaluation and assessment of XAI methods in LLM field.

\subsubsection{Issues}
\cite{bowman2023eight} critically examines LLMs, highlighting their unpredictability and the emergent nature of their capabilities with scaling. They underscore the challenges in steering and interpreting LLMs and the necessity for a nuanced understanding of their limitations and potential.

\cite{liu2023trustworthy} offers a survey and set of guidelines for assessing the alignment of LLMs with human values and intentions. They categorise and detail aspects of LLM trustworthiness, including reliability, safety, fairness, resistance to misuse, explainability, adherence to social norms, and robustness.

\cite{liao2023ai} emphasise the need for transparency in LLMs from a human-centred perspective. The authors discuss the unique challenges of achieving transparency with LLMs, differentiating them from smaller, more specialised models. The paper proposes a roadmap for research, emphasising the importance of understanding and addressing the transparency needs of diverse stakeholders in the LLM ecosystem. It advocates for developing and designing transparency approaches that consider these stakeholder needs, the novel applications of LLMs, and their various usage patterns and associated challenges.

Lastly,~\cite{xie2023wall} highlights the limitations of ChatGPT in explainability and stability in the context of financial market analysis through a zero-shot analysis. The authors suggest the need for more specialised training or fine-tuning.
\subsubsection{Benchmark and Metrics}


\cite{lu2022learn} introduce SCIENCEQA, a new dataset for multimodal science question answering. This dataset includes around 21k questions with diverse science topics and annotations, featuring lectures and explanations to aid in understanding the reasoning process. The authors demonstrate how language models, particularly LLMs, can be trained to generate these lectures and explanations as part of a CoT process, enhancing their reasoning capabilities. The study shows that CoT improves question-answering performance and provides insights into the potential of LLMs to mimic human-like multi-step reasoning in complex, multimodal domains.

\cite{golovneva2022roscoe} introduce ROSCOE, a set of metrics designed to evaluate the step-by-step reasoning of language models, especially in scenarios without a golden reference. This work includes a taxonomy of reasoning errors and a comprehensive evaluation of ROSCOE against baseline metrics across various reasoning tasks. The authors demonstrate ROSCOE's effectiveness in assessing semantic consistency, logicality, informativeness, fluency, and factuality in model-generated rationales.

\cite{zhao2023explainability} presents a comprehensive survey on explainability techniques for LLMs, focusing on Transformer-based models. It categorises these techniques based on traditional fine-tuning and prompting paradigms, detailing methods for generating local and global explanations. The paper addresses the challenges and potential directions for future research in explainability, highlighting LLMs' unique complexities and capabilities compared to conventional deep-learning models. Nevertheless, the survey mainly focuses on XAI in general and has minimal coverage of the relationship between XAI and LLMs.

\section{Discussion}
\label{sec:disscussion}
Our analysis indicates that a limited number of the reviewed publications directly tackle the challenges highlighted in Section \ref{sec:need}. For example, the work by~\cite{liu2023trustworthy} focuses on trust-related concerns in LLMs, whereas~\cite{gao2023chatgpt} investigates the issue of misinformation propagation by LLMs. This scant attention to the identified problems suggests an imperative for substantial engagement from the XAI community to confront these issues adequately.
\paragraph{Open-Source Engagement}
Our survey study shows that more studies are moving beyond the traditional approach of merely describing methodologies in text. Instead, they release them as tangible tools or open-source code, frequently hosted on platforms such as GitHub. This evolution is a commendable step toward enhancing transparency and reproducibility in computer science research. The trend suggests a growing inclination among authors to release their code and publicly publish their tools, a notable change from a few years ago. 
However, we should also mention the inconsistency in the level of community engagement with these repositories. While some repositories attract substantial interest, fostering further development and improvement, others remain underutilised. This disparity in engagement raises important questions about the factors influencing community interaction with these resources. 

\paragraph{Target}
Predominantly, most works have directed their attention towards LLMs rather than concentrating on more specialised or narrower subjects within AI-based systems. This broad approach contrasts the relatively few studies that focus specifically on Transformers or are confined to examining particular categories of systems, such as recommendation systems. This overarching focus on LLMs represents a positive and impactful trend within the AI community. Given the rapid development and increasing prominence of LLM systems in academic and practical applications, this broader focus is timely and crucial for driving our understanding and capabilities in this domain forward. It ensures that research keeps pace with the advancements in the field, fostering a comprehensive and forward-looking approach essential for AI technologies' continued growth and evolution.

\paragraph{Goal}
Our analysis, as delineated in Table~\ref{tab:studies}, reveals a bifurcation in the objectives of the LLM studies under review. On the one hand, a subset of these works is primarily dedicated to explaining and enhancing the interpretability of these 'black box' models. On the other hand, a larger contingent is more task-oriented, focusing on augmenting specific tasks and models, with interpretability emerging merely as a byproduct. This dichotomy in research focus underscores a pivotal trend: a pressing need to shift more attention towards demystifying the inner workings of LLMs. Rather than solely leveraging these models to boost task performance, their inherently opaque nature should not be overlooked. The pursuit of performance improvements must be balanced with efforts to unravel and clarify the underlying mechanisms of LLMs. This approach is crucial for fostering a deeper understanding of these complex systems, ensuring their application is effective and transparent. Such a balanced focus is essential for advancing the field technically and maintaining ethical and accountable AI development.
\section{Conclusion}
\label{sec:conclusion}

Our SMS reveals that only a handful of works are dedicated to developing explanation methods for LLM-based systems. This finding is particularly salient, considering the rapidly growing prominence of LLMs in various applications. Our study, therefore, serves a dual purpose in this context. Firstly, it acts as a navigational beacon for the XAI community, highlighting the fertile areas where efforts to create interpretable and transparent LLM-based systems can effectively address the challenges the broader AI community faces. Secondly, it is a call to action, urging researchers and practitioners to venture into this relatively underexplored domain. The need for explanation methods in LLM-based systems is not just a technical necessity but also a step towards responsible AI practice. By focusing on this area, the XAI community can contribute significantly to making AI systems more efficient, trustworthy and accountable.

Our \textbf{call for action} is as follows: Firstly, \textbf{researchers employing LLM models} must acknowledge and address the potential long-term challenges posed by the opacity of these systems. The importance of explainability should be elevated from a mere 'nice-to-have' feature to an integral aspect of the development process. This involves a proactive approach to incorporate explainability in the design and implementation phases of LLM-based systems. Such a shift in perspective is essential to ensure that these models are effective, transparent and accountable. Secondly, we urge \textbf{researchers in the XAI field} to broaden their investigative scope. The focus should not only be on devising methodologies capable of handling the complexity of LLM-based systems but also on enhancing the presentation layer of these explanations. Currently, explanations provided are often too complex for non-technical stakeholders. Therefore, developing approaches that render these explanations more accessible and understandable to a wider audience is imperative. This dual approach will make LLMs more understandable and user-friendly and bridge the gap between technical efficiency and ethical responsibility in AI development.


\bibliographystyle{unsrtnat}
\bibliography{bib}
\end{document}